\documentclass[runningheads]{llncs}
\usepackage{graphicx}
\usepackage{booktabs}
\usepackage{amsmath}
\DeclareMathOperator*{\argmin}{arg\,min}
\begin{document}
\title{Multi-Modality Information Fusion for Radiomics-based Neural Architecture Search\thanks{Supported by Australian Research Council (ARC) grants.}}
\titlerunning{Multi-Modality Neural Architecture Search}
%
\author{Yige Peng\inst{1} \and
Lei Bi\inst{1} \and
Michael Fulham\inst{1,2} \and
Dagan Feng\inst{1,3} \and
Jinman Kim\inst{1}}
%
\authorrunning{Y. Peng et al.}
%
\institute{School of Computer Science, University of Sydney, NSW, Australia\and
Department of Molecular Imaging, Royal Prince Alfred Hospital, NSW, Australia\and
Med-X Research Institute, Shanghai Jiao Tong University, Shanghai, China\\}
\maketitle              
\begin{abstract}
‘Radiomics’ is a method that extracts mineable quantitative features from radiographic images. These features can then be used to determine prognosis, for example, predicting the development of distant metastases (DM). Existing radiomics methods, however, require complex manual effort including the design of hand-crafted radiomic features and their extraction and selection. Recent radiomics methods, based on convolutional neural networks (CNNs), also require manual input in network architecture design and hyper-parameter tuning. Radiomic complexity is further compounded when there are multiple imaging modalities, for example, combined positron emission tomography – computed tomography (PET-CT) where there is functional information from PET and complementary anatomical localization information from computed tomography (CT). Existing multi-modality radiomics methods manually fuse the data that are extracted separately. Reliance on manual fusion often results in sub-optimal fusion because they are dependent on an ‘expert’s’ understanding of medical images. In this study, we propose a multi-modality neural architecture search method (MM-NAS) to automatically derive optimal multi-modality image features for radiomics and thus negate the dependence on a manual process. We evaluated our MM-NAS on the ability to predict DM using a public PET-CT dataset of patients with soft-tissue sarcomas (STSs). Our results show that our MM-NAS had a higher prediction accuracy when compared to state-of-the-art radiomics methods.

\end{abstract}
\section{Introduction}
The implementation and availability of high-throughput computing has made it possible to extract innumerable features from medical imaging datasets. These extracted features can reveal disease related characteristics that can relate to prognosis \cite{lambin2012radiomics}. The process of converting visual imaging data into mineable quantitative features is referred to radiomics \cite{hatt2019radiomics}. Radiomics is an emerging field of translational research in medical imaging where the modalities include digital radiography, magnetic resonance imaging (MRI), computed tomography (CT), combined positron emission tomography – computed tomography (PET-CT) etc. The range of medical imaging modalities is wide and, in essence, these modalities provide information about structure, physiology, pathology, biochemistry and pathophysiology \cite{rizzo2018radiomics}. PET-CT, for example, combines the sensitivity of PET in detecting regions of abnormal function and the specificity of CT in depicting the underlying anatomy of where the abnormal functions are occurring. Multi-modality PET-CT, therefore, is regarded as the imaging modality of choice for the diagnosis, staging and monitoring the treatment response of many cancers \cite{hatt2017characterization}.  Conventional radiomics studies mainly focus on encoding regions of interest (e.g., tumors), with hand-crafted features, such as intensity, texture, shape, etc. These features are used to build conventional predictive models such as multivariable statistical analysis \cite{vallieres2015radiomics}, support vector machine (SVM) \cite{juntu2010machine} and random forest \cite{vallieres2017radiomics}. Unfortunately, these methods rely on prior knowledge in hand-crafting image features and tuning of a large number of parameters for building the predictive models.

Radiomics methods based on convolutional neural networks (CNN) are regarded as the state-of-the-art because they can learn high-level semantic image information in an end-to-end fashion. CNN-based radiomics methods were mainly designed for 2D single-modality images such as CT \cite{kumar2017discovery,diamant2019deep} and MRI \cite{zhu2019deep}. For the limited methods that attempted to fuse multi-modality images, the focus was on fusing the image features that were separately extracted from the individual modalities \cite{lao2017deep,li2017deep,peng2019deep}. In addition, these methods required human expertise to design the dataset specific architectures e.g., the number of convolutional layers, the layer to fuse multi-modality image features. Architecture design and optimization require a large amount of domain knowledge such as in validating the architecture performance and tuning the hyper-parameters. Neural architecture search (NAS) has recently been proposed to simplify the challenges in architecture design by automatically searching for an optimal net-work architecture based on a given dataset. The NAS thus enables reduced manual input and reliance on prior knowledge \cite{elsken2018neural}. Investigators have attempted to apply the NAS for single medical imaging modality related tasks but the main focus has been on segmentation \cite{10.1007/978-3-030-32245-8_26,10.1007/978-3-030-32226-7_92}.

We propose a multi-modality NAS method (MM-NAS) to search for a multi-modality CNN architecture for use in PET-CT radiomics. Our contribution, when compared to existing methods includes: (i) the ability to build an optimal, fully-automated radiomics CNN architecture; (ii) enabling an optimal fusion of PET-CT images for radiomics. Our method finds various fusion modules e.g., fusion via different network operations (e.g., convolution, pooling, etc.) at different stages of the network. These searched fusion modules provide more options for integrating the complementary PET and CT data. We outline how our approach can predict the development of distant metastases (DM) in patients with soft-tissue sarcomas (STSs). STSs include slow-growing, more well-differentiated tumors, aggressive tumors that grow rapidly and spread to other organs (distant metastases – DM) and more intermediate that behave between the two extremes \cite{billingsley1999multifactorial,yachida2010distant}. The early identification of patient who may develop metastatic disease may contribute to improved care and better patient outcomes.

\section{Method}
\subsection{Materials}
We used a public PET-CT STSs dataset from the cancer imaging archive (TCIA) repository \cite{vallieres2015radiomics,clark2013cancer}. The dataset has 51 multi-modality PET-CT scans derived from 51 patients with pathology-proven STSs. DM were confirmed via biopsy or diagnosed by an expert clinician. Three patients without clear metastases information was excluded. Thus, our dataset consists of 48 studies, half of which developed DM.

\subsection{Neural Architecture Search Setting}
\begin{figure}
\centering
\includegraphics[width=\textwidth]{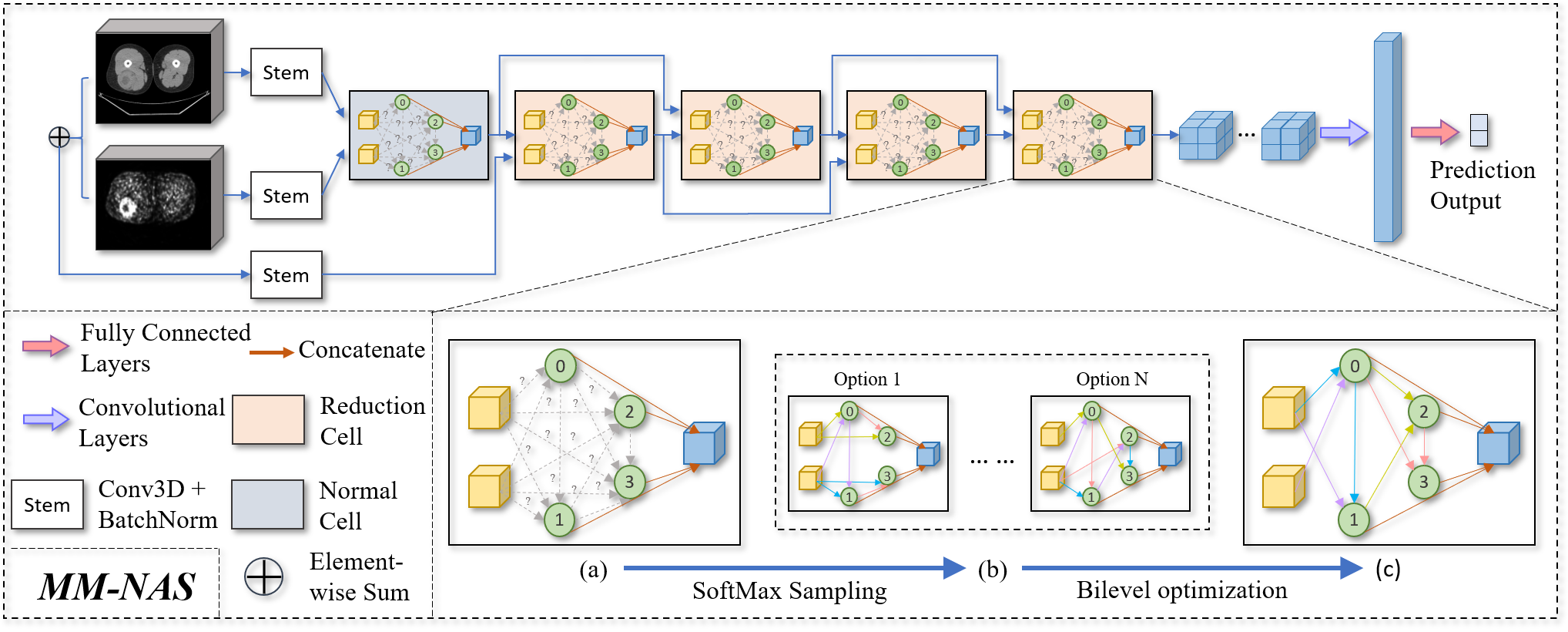}
\caption{MM-NAS overview – the CNN architecture has multiple different cells (normal, reduction); each cell is a directed acyclic graph as the basic unit; directed arrows indicate the forward path: (a) initial operations on edges of each cell are unknown; (b) continuously production of alternative cells by SoftMax sampling; and (c) optimal cell architecture after iterative bilevel optimization.} \label{fig1}
\end{figure}

We followed the existing NAS methods \cite{zoph2018learning,liu2018darts,pham2018efficient} and focused on searching of different computational cells (normal, reduction) to improve the computational efficiency. The computational cells are the basic unit that can be stacked multiple times to form a CNN. A NAS workflow is as follows: (i) based on the given training data, search for optimal cell structure that can form a CNN; and (ii) train the searched CNN based on the training data and then evaluate on the testing data. In our MM-NAS (as shown in Fig.~\ref{fig1}), every cell is regarded as a directed acyclic graph consisting of two inputs, one output and several ordered nodes. Our MM-NAS has normal and reduction cells. The input and the output feature maps of a normal cell have the same dimensions. The reduction cell doubles the channel number and reduces the input feature map by half. A stem block consists of a 3D convolutional layer and a batch normalization layer and is used for input image transitions. In our method, the outputs of PET and CT stem blocks are separately fed into the first normal cell to facilitate the fusion process. Then the output feature maps of the first normal cell flows into the first reduction cell with the sum of PET and CT image, which is also processed by one stem block. The rest of the reduction cells used the output feature maps from the previous two layers as input. For DM prediction, the output feature maps of last reduction cell were fed into two convolutional layers and one fully connected layer for classification.

\subsection{Optimization Strategy}
Each intermediate node $n^i$ inside a cell is a feature map. We represent the searched operations on edge $(i,j)$ using the vector 
$x^{(i, j)}\left(n^{j}\right)=\left\{x_{\sigma}^{(i, j)} \mid \sigma \in O\right\}$
and the vector of all optional operations as 
$\mathbf{0}^{(i, j)}=\left\{\sigma\left(n^{i} ; \vartheta_{\sigma}^{(i, j)}\right) \mid \sigma \in O\right\}$, 
where $O$ denotes the set of optional operations, $\vartheta_{\sigma}^{(i, j)}$ denotes the parameters of the operation $\sigma$ on edge $(i,j)$. Then the intermediate nodes can be computed by sum of all their predecessors:
\begin{equation}
n^{j}=\left\langle x^{(i, j)}\left(n^{j}\right), \mathbf{0}^{(i, j)}\right\rangle=\sum_{i<j, \sigma \in O} x^{(i, j)}\left(n^{j}\right) \sigma\left(n^{i} ; \vartheta_{\sigma}^{(i, j)}\right)
\end{equation}
\noindent As the possible operations are mixed through a SoftMax function, this makes the search space continuous:
\begin{equation}
x^{(i, j)}\left(n^{j}\right)=\sum_{\sigma \in O} \frac{\exp \left(\alpha_{\sigma}^{(i, j)}\right)}{\sum_{\sigma^{\prime} \in O} \exp \left(\alpha_{\sigma^{\prime}}^{(i, j)}\right)} x_{\sigma}^{(i, j)}
\end{equation}
\noindent Where $\alpha_\sigma^{\left(i,\ j\right)} $denotes a probability distribution over the operation set $O$.

Denote by $L_{train}$ and $L_{val}$ the training and the validation loss. Because both losses are determined not only by the architecture $\alpha$ , but also the weights $\theta$ in the network, where
$\theta=\left\{\left(\vartheta^{\left(i,\ j\right)}\right)\middle|\left(i,\ j\right)\in C\right\}$, 
$C$ is the computational cell. The aim of searching the best architecture is to find a proper $\alpha$ that minimizes the validation loss $L_{val}\left(\theta^\ast\left(\alpha\right),\alpha\right)$, where the weights $\theta$ associated with the architecture are obtained by minimizing the training loss:
\begin{equation}
\min_\alpha{\ L_{val}\left(\theta^\ast\left(\alpha\right),\alpha\right)}
\end{equation}
\begin{equation}
s.t.\ \ \theta^\ast\left(\alpha\right)=\argmin_\theta{L_{train}\left(\theta,\alpha\right)}
\end{equation}

\subsection{Implementation Details}
We implemented our MMR-NAS in PyTorch. The input image size was fixed to 112$\times$112$\times$144. The operation set $\mathbf{O}^{\left(i,\ j\right)}$ for each cell includes 3D standard convolutions, 3D separable convolutions, 3D dilated convolutions, 3D max pooling, 3D average pooling, skip connections and zero operations. All operations are of stride one (if applicable) and the kernel size of pooling operations are 3. The kernel size for the convolutional operations can either be 3 or 5. Cross-entropy loss was used during the architecture search step for training optimization. The parameters of each cell were optimized by Adam with a learning rate of 0.0005 while the weight in the whole network was optimized by SGD with a learning rate of 0.0001, and the batch size was set to 1. It took about 3 minutes to process one epoch with 40 PET-CT volumetric training images, and the best architecture was obtained at epoch 70 out of total 200 epochs. Cross-entropy loss with Adam was used for training optimization in the second step for training the searched architecture. Learning rate was set to with 0.001 and batch size was set to 1. It took 2 minutes to train one epoch, the best model was obtained at approximately epoch 80 out of 200 epochs. All the experiments were conducted on a 11GB NVIDIA GeForce GTX 2080Ti GPU.

\subsection{Experimental Setup}
We conducted the following experiments: (a) a comparison with the state-of-the-art radiomics methods; (b) compared the performance of using multi-modality CNNs to single-modality CNNs; and (c) compared the performance of using 2D CNNs with 3D CNNs for radiomics. In experiment (a), we compared our MM-NAS with the following methods: (i) HC+RF – we followed the conventional radiomics method \cite{vallieres2017radiomics} used hand-crafted (HC) features (e.g. intensity solidity, skewness, grey-level co-occurrence matrix features, etc.) extracted from tumor region with random forest (RF) as the classifier for predication; (ii) DLHN – a deep learning based  head \& neck cancer outcome (e.g., DM, loco-regional failure, and overall survival) prediction \cite{diamant2019deep}; (iii) 3DMCL – a deep learning based 3D based multi-modality collaborative learning for distant metastases prediction with PET-CT images \cite{peng2019deep}. We used a 6-fold cross-validation approach for the MM-NAS and the comparison methods. In each-fold cross-validation, we used 40 PET-CT images for training and the remaining 8 images for testing. Six well established evaluation metrics were used for comparison including accuracy (acc.), precision (pre.), F1 score (F1) and area under the receiver-operating characteristic curve (AUC).

\section{Results}
\begin{figure}
\centering
\includegraphics[width=\textwidth]{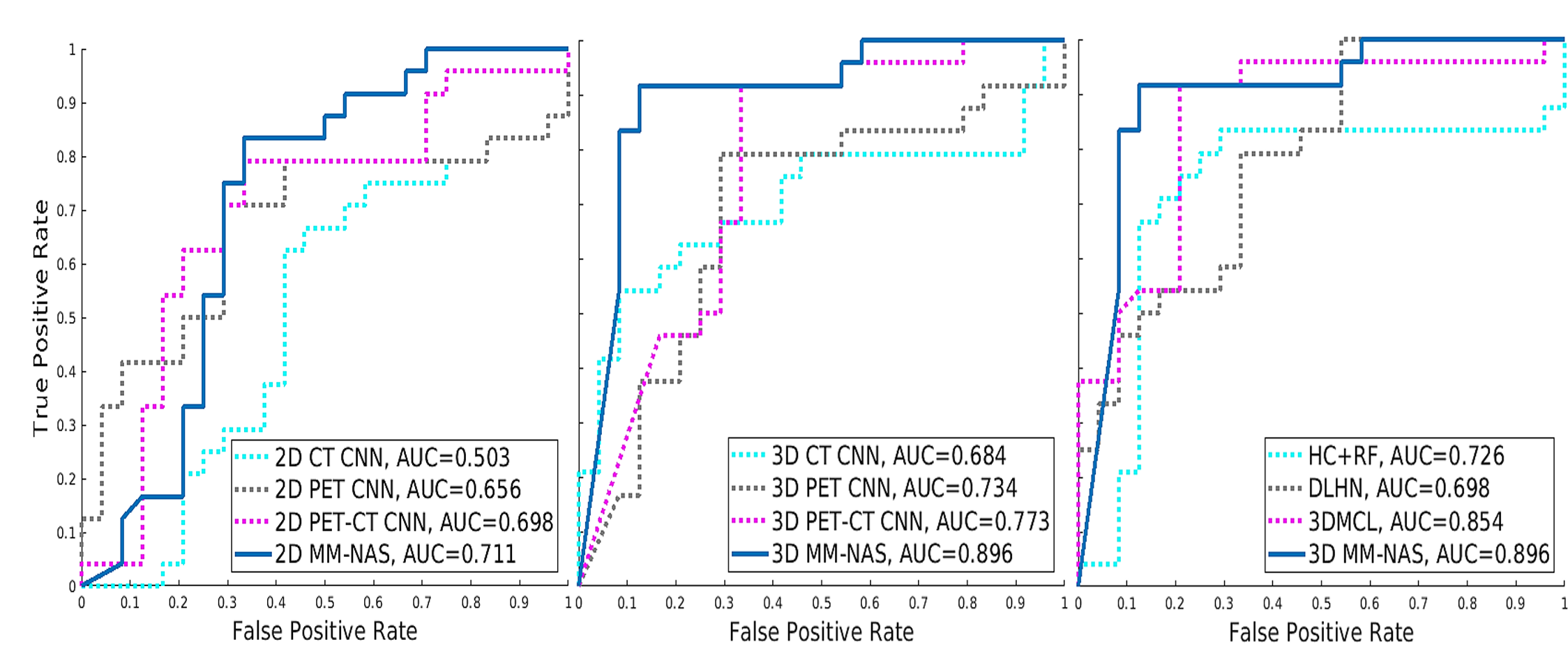}
\caption{ROC curves of ours and comparative radiomics methods.} \label{fig2}
\end{figure}

The receiver-operating characteristic (ROC) curve is shown in Fig.~\ref{fig2}. It shows that our 2D MM-NAS achieved better performance when compared with 2D CNN based methods. Our 3D MM-NAS outperformed other 3D CNN based comparison methods and achieved the overall best performance.

Table ~\ref{tab1} and Table ~\ref{tab2} present results of 3D MM-NAS achieving the best outcomes in all measures with AUC value of 0.896, accuracy of 0.896, sensitivity of 0.917, specificity of 0.875, precision of 0.880, and F1 score of 0. 898. 

\begin{table}
\centering
\caption{Comparison of state-of-the-art radiomics methods.}\label{tab1}
\begin{tabular*}{0.9\textwidth}{@{\extracolsep{\fill}}lcccccc@{}}
\toprule
Evaluation Metrics & Acc. & Sen. & Spe.	& Pre. & F1. & AUC\\
\midrule
HC+RF \cite{vallieres2017radiomics} & 0.750	& 0.792	& 0.708	& 0.731	& 0.760	& 0.726\\
DLHN \cite{diamant2019deep} & 0.729	& 0.792	& 0.667	& 0.703	& 0.745	& 0.698\\
3DMCL \cite{peng2019deep} & 0.854 & 0.917 & 0.792 & 0.815 & 0.863 & 0.854\\
2D MM-NAS (Ours) & 0.750 & 0.833 & 0.667 & 0.714 & 0.769 & 0.711\\
3D MM-NAS (Ours) & \textbf{0.896} & \textbf{0.917} & \textbf{0.875} & \textbf{0.880} & \textbf{0.898} & \textbf{0.896}\\
\bottomrule
\end{tabular*}
\end{table}

\begin{table}
\centering
\caption{Comparison of methods using different imaging modalities with convolutional kernels.}\label{tab2}
\begin{tabular*}{0.9\textwidth}{@{\extracolsep{\fill}}lcccccc@{}}
\toprule
Evaluation Metrics & Acc. & Sen. & Spe.	& Pre. & F1. & AUC\\
\midrule
2D CT CNN & 0.583 & 0.708 & 0.458 & 0.567 & 0.630 & 0.503\\
2D PET CNN & 0.729 & 0.542 & \textbf{0.917} & \textbf{0.867} & 0.667 & 0.656\\
2D PET-CT CNN & 0.729 & 0.792 & 0.667 & 0.703 & 0.745 & 0.698\\
2D MM-NAS (Ours) & \textbf{0.750} & \textbf{0.833} & 0.667 & 0.714 & \textbf{0.769} & \textbf{0.711}\\
\midrule
3D CT CNN & 0.667 & 0.667 & 0.667 & 0.667 & 0.667 & 0.684\\
3D PET CNN & 0.771 & 0.750 & 0.792 & 0.783 & 0.766 & 0.734\\
3D PET-CT CNN & 0.792 & 0.792 & 0.792 & 0.792 & 0.792 & 0.773\\
3D MM-NAS (Ours) & \textbf{0.896} & \textbf{0.917} & \textbf{0.875} & \textbf{0.880} & \textbf{0.898} & \textbf{0.896}\\
\bottomrule
\end{tabular*}
\end{table}

\section{Discussion}
Our main findings are that our MM-NAS: (i) performs better than the commonly used radiomics methods and, (ii) derives optimal multi-modality radiomic features from PET-CT images; (iii) removes the reliance on prior knowledge when building the optimal CNN architecture.

\begin{figure}
\centering
\includegraphics[width=\textwidth]{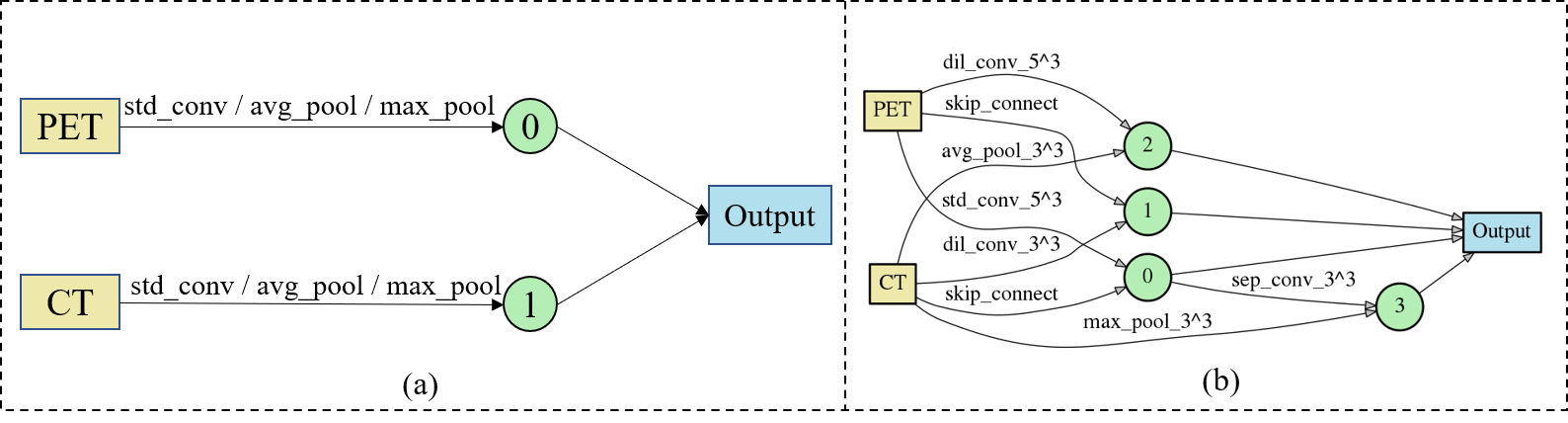}
\caption{The comparison between (a) the simplified fusion approach of existing approaches, such as DLHN and 3DMCL; (b) the learned normal cell of MM-NAS for PET-CT fusion.} \label{fig3}
\end{figure}

We attribute the improved performance of our MM-NAS to the search of the optimal computation cells, within the NAS, that allowed for fusing multi-modality image features at different stages of the network. Existing approaches often choose to fuse the separately extracted feature maps after several convolutional / pooling layers (see Fig.~\ref{fig3}). Our derives cell structure offers more freedom to integrate multi-modality images via various operations and connections, thus producing the optimal radiomic features to predict distant disease. The state-of-the-art method 3DMCL outperformed HC+RF and DLHN due to the collaborative learning of both pre-defined radiomic features and deep features, whereas our MM-NAS obtained better performance over all the evaluation metrics without feature handcrafting. Thus, the elimination of prior knowledge could contribute to a better generalizability for applications in other radiomics studies.

The differences between PET-CT CNN and CNN with PET or CT alone show the advantage of incorporating multi-modality information. Across the single modality CNNs, PET-based methods outperformed CT-based methods. We ascribe this to the functional features, which can better characterize the tumor, when compared to anatomical features from CT that rely on changes in size which are often a later development. Such features from PET could potentially uncover functional information that relate to the biological behavior of tumors \cite{hatt2017characterization}.

The relatively poor performance of 2D CNNs when compared to 3D CNNs is expected. This is attributed to the fact that volumetric image features derived from 3D CNNs are better able to derive spatial information e.g., volumetric tumor shape and size. Spatial information has strong correlations to the DM predictions \cite{hosny2018deep}.

\section{Conclusion}
We have outlined a multi-modality neural architecture search method (MM-NAS) for PET-CT to predict the development of distant disease (metastases) in patient with STSs. Our method automatically searched for a multi-modality CNN based radiomics architecture and the architecture can then be used to fuse and derive the optimal PET-CT image features. Our results show that our PET-CT image features are the most relevant for predicting distant metastases.

%
%
%
\bibliographystyle{splncs04}
\bibliography{mybibliography}

\end{document}